\documentclass[conference]{IEEEtran}
\IEEEoverridecommandlockouts
\usepackage{cite}
\usepackage{amsmath,amssymb,amsfonts}
\usepackage{algorithmic}
\usepackage{graphicx}
\usepackage{textcomp}
\usepackage{xcolor}
\def\BibTeX{{\rm B\kern-.05em{\sc i\kern-.025em b}\kern-.08em
    T\kern-.1667em\lower.7ex\hbox{E}\kern-.125emX}}

\usepackage{titlesec}
\makeatletter
\@addtoreset{subsubsection}{subsection}
\makeatother

\titleformat{\subsubsection}[block]
  {\normalfont\normalsize}%
  {\Alph{subsection}.\arabic{subsubsection})}
  {0.5em}{}%
\titlespacing*{\subsubsection}{0pt}{1ex}{0.5ex}

\usepackage{booktabs} 

\renewcommand{\baselinestretch}{0.97}

\usepackage[hidelinks]{hyperref}

\usepackage{cleveref}
\Crefname{equation}{Eq.}{Eqs.} 
\Crefname{section}{Sec.}{Secs.}
\Crefname{figure}{Fig.}{Fig.}
\Crefname{table}{Tab.}{tabs.}

\usepackage[inkscapearea=page]{svg}

\usepackage{tcolorbox}
\tcbuselibrary{theorems}

\usepackage{balance}

\usepackage{siunitx}

\usepackage{bm}

\usepackage{graphicx}

\usepackage{enumitem}

\usepackage{mathtools}

\usepackage{tikz}
\usetikzlibrary{shapes.geometric, arrows}
\usetikzlibrary{external}
\tikzstyle{startstop} = [rectangle, minimum width=0.48\textwidth, minimum height=0.5cm,text centered, draw=black, fill=gray!20]
\tikzstyle{arrow} = [thick,->,>=stealth]

\begin{document}

\bstctlcite{bibcontrol_etal4}

\title{Dynamics-Decoupled Trajectory Alignment \\for Sim-to-Real Transfer in Reinforcement Learning \\for Autonomous Driving
\thanks{This research paper [project MORE] is funded by dtec.bw -- Digitalization and Technology Research Center of the Bundeswehr. dtec.bw is funded by the European Union -- NextGenerationEU.  

All authors are with the Chair of Machine Perception for Autonomous
Driving, University of the Bundeswehr Munich, Germany. Contact author email:
        {\tt\small
        thomas.steinecker@unibw.de
        }
}
}

\author{Thomas Steinecker, Alexander Bienemann, Denis Trescher, Thorsten Luettel and  Mirko Maehlisch}

\maketitle


\begin{abstract}

Reinforcement learning (RL) has shown promise in robotics, but deploying RL on real vehicles remains challenging due to the complexity of vehicle dynamics and the mismatch between simulation and reality. Factors such as tire characteristics, road surface conditions, aerodynamic disturbances, and vehicle load make it infeasible to model real-world dynamics accurately, which hinders direct transfer of RL agents trained in simulation.
In this paper, we present a framework that decouples motion planning from vehicle control through a spatial and temporal alignment strategy between a virtual vehicle and the real system. An RL agent is first trained in simulation using a kinematic bicycle model to output continuous control actions. Its behavior is then distilled into a trajectory-predicting agent that generates finite-horizon ego-vehicle trajectories, enabling synchronization between virtual and real vehicles. At deployment, a Stanley controller governs lateral dynamics, while longitudinal alignment is maintained through adaptive update mechanisms that compensate for deviations between virtual and real trajectories.
We validate our approach on a real vehicle and demonstrate that the proposed alignment strategy enables robust zero-shot transfer of RL-based motion planning from simulation to reality, successfully decoupling high-level trajectory generation from low-level vehicle control.

\end{abstract}




\section{Introduction}
\label{sec:introduction}

Reinforcement learning (RL) has become increasingly popular for solving complex decision-making tasks, with notable successes such as AlphaZero in board games~\cite{silver2017mastering}, autonomous drone racing~\cite{kaufmann2023champion}, and protein structure prediction~\cite{abramson2024accurate}. Its potential for autonomous driving is widely recognized, but most research remains limited to simulation or open-loop evaluation \cite{teng2023motion}, and the sim-to-real gap persists as a major challenge. A central reason for this gap is the difficulty of capturing the full complexity of real-world vehicle dynamics.

Vehicle dynamics are highly sensitive to a variety of operational and environmental parameters such as tire temperature, inflation pressure, road surface type, vehicle load, and steering/slip behavior. For example, lateral tire dynamics (cornering stiffness, relaxation length, friction) vary significantly with tire temperature, vertical load, and inflation pressure~\cite{alcazar2022modeling}. Accurately modeling all of these effects is infeasible, which creates a fundamental challenge for transferring reinforcement learning (RL) policies trained in simulation to real-world driving. As a result, discrepancies between simplified training models and real vehicle dynamics commonly degrade the transfer performance of learned policies when deployed on physical systems~\cite{zhao2020sim,voogd2023digitaltwin}.

\begin{figure}
    \centering
    \includegraphics[width=0.94\columnwidth]{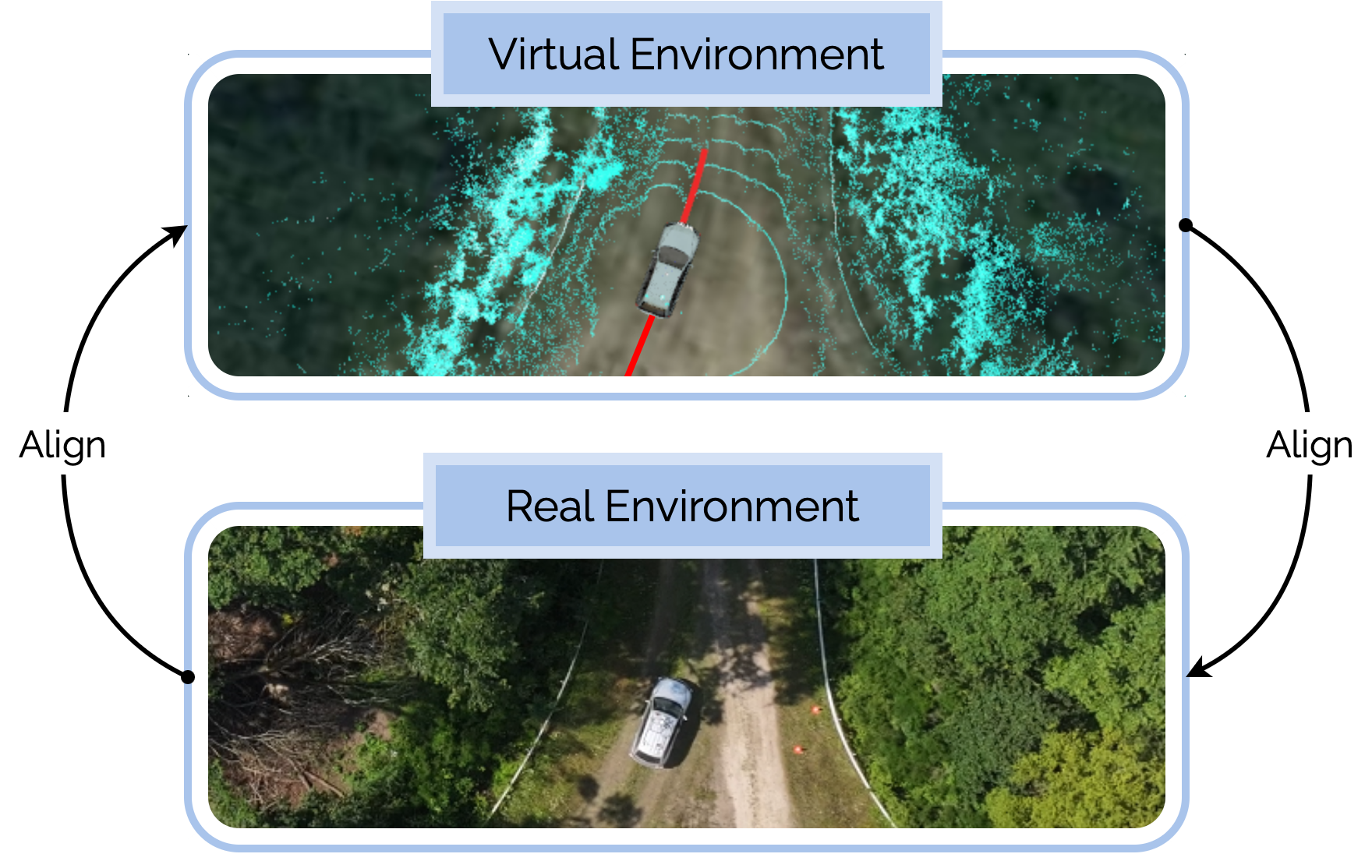}
    \caption{Concept of virtual-to-real trajectory alignment. A kinematic vehicle model, initialized with the real state, generates a virtual trajectory. Alignment strategies ensure that the real vehicle remains longitudinally and laterally synchronized with this virtual reference.}
    \label{fig:overview}
\end{figure}

To address this challenge, we propose to decouple the dynamics of the virtual training environment from those of the real vehicle. By introducing a trajectory-predicting agent and aligning its virtual execution with the real system, our framework enables safe and robust transfer of RL policies from simulation to reality with respect to dynamics in a zero-shot manner.

Our approach begins with training an RL agent in simulation using a kinematic bicycle model. The agent is optimized to follow noisy paths while maintaining a target velocity and adhering to safety-related constraints such as bounded centripetal force. The agent directly outputs instantaneous low-level control commands (acceleration and steering rate) and is trained with a reward function that balances tracking accuracy, driving smoothness, and constraint satisfaction.

As a next step, we distill the learned control behavior into a trajectory-generating agent. Instead of predicting only the next action, this agent outputs an entire sequence of future actions over a finite horizon. During training, only the first action of this trajectory is applied to the virtual vehicle, while the remaining predictions serve as additional supervisory signals. This trajectory representation not only provides a richer description of intended motion but also enables subsequent alignment with the real vehicle.

At deployment, the real vehicle is synchronized with a virtual counterpart initialized to the same state. The virtual vehicle evolves according to the kinematic model and the first action of the trajectory-predicting agent. Its predicted trajectory serves as the desired reference for a Stanley controller \cite{hoffmann2007autonomous}, which governs the lateral dynamics of the real vehicle. The use of autoregressive prediction with injected action noise further improves robustness to distributional shift. Longitudinal alignment between the real and virtual vehicles is enforced through a feed-forward and feed-back control. If the real vehicle falls behind or moves ahead of admissible bounds, the framework adjusts by either applying no update step at all (to let the real vehicle catch up) or apply two update steps at once (to let the virtual vehicle catch up). This ensures that the real vehicle tracks the intended path while maintaining safety and robustness.

Through this combination of reinforcement learning, trajectory-based knowledge distillation, and dynamic alignment between real and virtual vehicles, we provide a principled approach to transferring learned control strategies from simulation to real-world autonomous driving.


Several works have explored the application of RL to real vehicles. One of the earliest demonstrations is \cite{folkers2019controlling}, where RL agents were shown to control an autonomous car at low speeds. The agent directly produced angular velocities and accelerations, but only on a step-by-step basis without any prediction horizon or explicit mechanism to mitigate dynamics mismatch. At the low speeds considered, dynamics discrepancies were less critical, which limited the generality of the approach.  

In \cite{kendall2019learning}, training was performed directly on a real vehicle using visual inputs for lane following. While this demonstrated the feasibility of training policies without simulation, the process is constrained by real-time data collection, is limited to relatively simple tasks, and does not scale to more complex environments such as urban driving.  

More recent efforts have investigated sim-to-real transfer. In \cite{maramotti2022tackling}, an additional dynamics model was trained alongside the policy to account for vehicle behavior under real-world conditions. However, such dynamics models are typically vehicle-specific, difficult to validate across operating conditions, and sensitive to variations in mass, tire properties, and surface conditions. Moreover, the agent directly predicted steering accelerations and steering angles, with the latter potentially leading to discontinuous or infeasible control inputs.

A different direction was taken in \cite{voogd2023digitaltwin}, which employed a high-fidelity simulator as a digital twin of the real vehicle. While this reduces the sim-to-real gap, it requires significant engineering effort and computational resources to construct and maintain the digital twin, and it still relies on accurate dynamics modeling that may not generalize to all operating conditions.  

For high-performance maneuvers, \cite{zhao2024high} combined RL with model predictive control (MPC) for aggressive cornering and drift. While this work successfully demonstrated RL-assisted control in challenging scenarios, the MPC was only used to track immediate actions or pre-specified trajectories, without an explicit mechanism to align longer-horizon predictions between simulation and reality. Moreover, precise vehicle calibration and high-fidelity dynamics models were essential, which limits applicability in less controlled settings.  

Other works have explored scaled platforms such as \cite{RLPP2024}, which applied residual RL to autonomous racing on small robotic platforms. While these platforms are valuable for validating algorithms, their dynamics are simpler and results do not directly transfer to full-scale vehicles. Finally, some approaches have focused primarily on the visual domain, such as \cite{osinski2020simulation}, which addressed the perception gap but relied on rudimentary control (constant velocity and steering-only decisions).  

Across these studies, two main patterns emerge: first, most existing works predict a single-step control action without providing trajectories that enable alignment between virtual predictions and real vehicle behavior; second, efforts to bridge the sim-to-real gap largely rely on improved simulators, domain randomization, or direct dynamics modeling. In contrast, our approach distills an RL policy into a trajectory-predicting agent and explicitly aligns a virtual vehicle with the real system. This decoupling of motion planning and control, combined with the Stanley controller for lateral dynamics and adaptive longitudinal synchronization, enables robust zero-shot, sim-to-real transfer without relying on high-fidelity simulators or vehicle-specific dynamics models.

The paper is structured as follows. \Cref{sec:methods} presents the proposed approach. \Cref{sec:evaluation} evaluates the alignment strategy on a real vehicle and discusses its potentials and limitations, and \Cref{sec:conclusion} concludes the work.


\section{Methods}
\label{sec:methods}

\Cref{sec:methods_rl} presents the reinforcement learning (RL) and modeling, whose results are used in \Cref{sec:methods_trajectory_generation} to derive a trajectory-generating agent via knowledge distillation. Finally, \Cref{sec:methods_alignment} describes how this agent is applied for virtual-to-real trajectory alignment.

\subsection{Reinforcement Learning in Simulation}
\label{sec:methods_rl}

The objective of the agent is to follow a noisy path while optimizing for desired velocity, tracking accuracy, control smoothness, and comfort, the latter quantified through centripetal force and changes in consecutive actions. 

For training, we use TD3~\cite{fujimoto2018addressing}, which supports continuous action spaces and operates off-policy, allowing the future use of open-loop data. To improve training efficiency, we normalize all inputs to a standard Gaussian distribution $\mathcal{N}(0, 1)$, employ a prioritized replay buffer~\cite{schaul2015prioritized}, and use n-step returns~\cite{sutton1998reinforcement} for better credit assignment.  

We now describe the simulation environment, policy representation, and reward design.  

\textbf{Environment Modelling}—The simulation environment is a lightweight kinematic-based simulator that generates random noisy paths. The vehicle is initialized at the start of each path with randomized deviations in lateral offset, heading, steering angle, and velocity to improve robustness.  

Path generation follows an Ornstein–Uhlenbeck (OU) process for curvature variation, ensuring feasible yet stochastic trajectories. The path starts with two points, and curvature is iteratively updated using an OU process before being clamped within admissible bounds. Each new position is computed by integrating the curvature into the heading update, yielding a sequence of waypoints. Additionally, to each waypoint Gaussian noise is added with $\mathcal{N}(0, 0.1)$, to robustify the path following training. 

\begin{figure}
    \centering
    \includegraphics[width=\columnwidth]{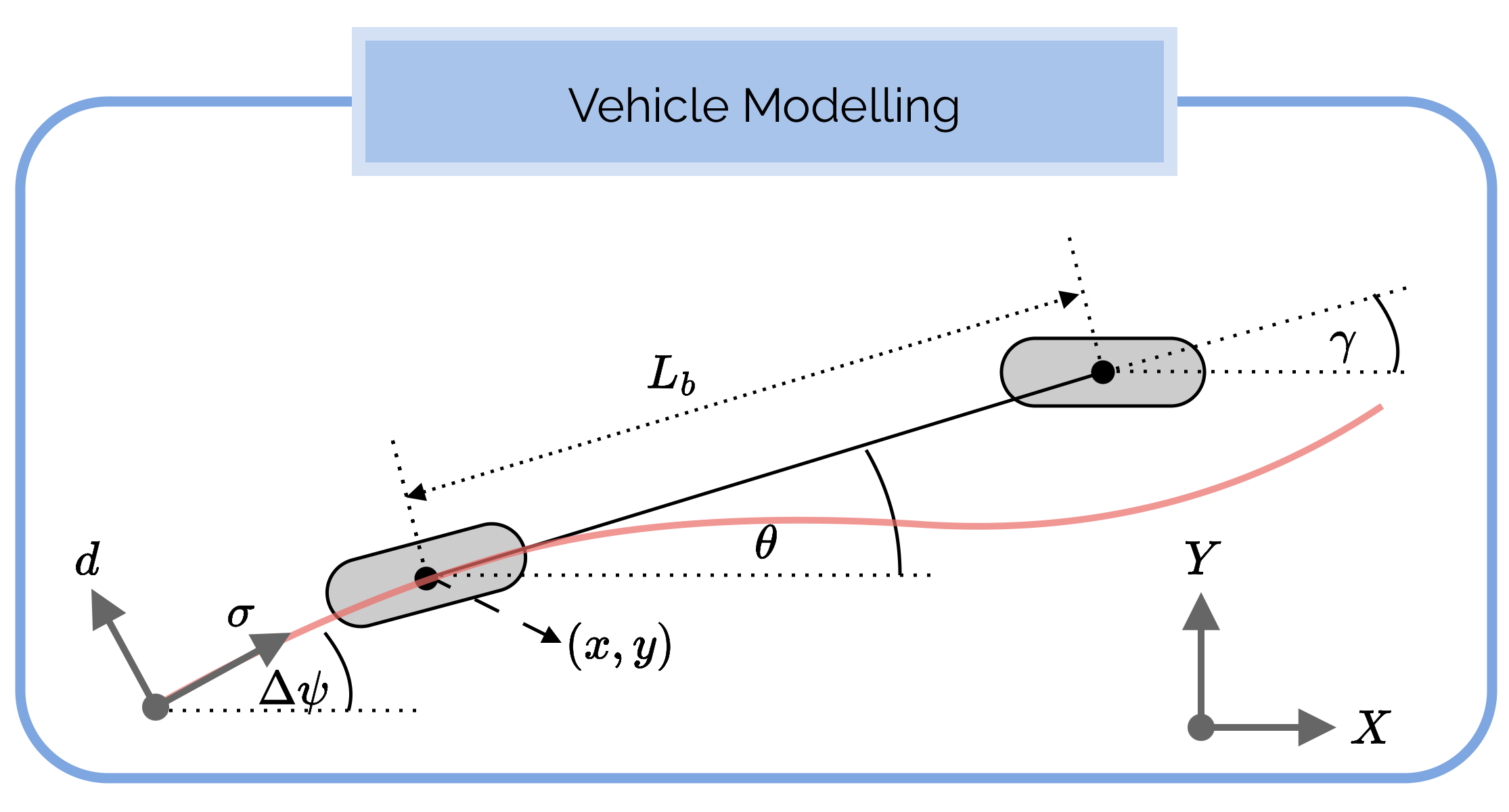}
    \caption{Bicycle model used in simulation. The inertial frame $(X, Y)$ is shown in the bottom right. The vehicle reference point $(x, y)$ lies at the rear axle center, with heading $\theta$, steering angle $\gamma$, and wheelbase $L_b$. The curvilinear coordinate frame is shown in the bottom left, with an exemplary path depicted in red.}
    \label{fig:vehicle_modelling}
\end{figure}

The vehicle dynamics are modeled with a kinematic bicycle model (\Cref{fig:vehicle_modelling}), where $(x,y)$ denotes the rear-axle reference point, $L_b$ the wheelbase, $\theta$ the heading, and $\gamma$ the steering angle. The discrete-time dynamics with step size $\Delta t$ are given by:
\begin{equation}
\begin{aligned}
\begin{bmatrix}
x_{t+1} \\[4pt]
y_{t+1} \\[4pt]
\theta_{t+1} \\[4pt]
v_{t+1} \\[4pt]
\gamma_{t+1}
\end{bmatrix}
=
\begin{bmatrix}
x_t + \Delta t \, v_t \cos(\theta_t) \\[4pt]
y_t + \Delta t \, v_t \sin(\theta_t) \\[4pt]
\theta_t + \Delta t \, \tfrac{v_t}{L_b} \tan(\gamma_t) \\[4pt]
v_t + \Delta t \, a_t \\[4pt]
\gamma_t + \Delta t \, \omega_t
\end{bmatrix},
\end{aligned}
\label{eq:bicycle_model}
\end{equation}
where $a_t$ is the longitudinal acceleration and $\omega_t$ the steering rate at time step $t$.

\textbf{Policy Representation}—The agent’s state at time $t$ is given by the concatenation of the observation $\mathcal{O}_t$ and the parameterization $\mathcal{P}_t$:  
\begin{equation}
s_t = \big[ \mathcal{O}_t, \; \mathcal{P}_t \big].
\end{equation}

The observation $\mathcal{O}_t$ includes the vehicle state and previous action, i.e.  
\begin{equation}
\mathcal{O}_t = \big[ x_t, \, y_t, \, \theta_t, \, v_t, \gamma_t, \, a_{t-1}, \, \omega_{t-1} \big],
\end{equation}

The parameterization $\mathcal{P}_t$ consists of the target velocity and $N_w$ waypoints ahead of the vehicle, sampled at approximately constant arc length $d_s$ along the path and expressed in vehicle coordinates:  
\begin{equation}
\mathcal{P}_t = \big[ v_{\text{target,t}}, \; \{ w_t^{(1)}, \ldots, w_t^{(N_w)} \} \big],
\end{equation}
with each waypoint $w_t^{(i)} = (x_t^{(i)}, y_t^{(i)})$.  

The action space $\mathcal{A}$ consists of the longitudinal acceleration $a_t$ and steering rate $\omega_t$, both bounded by conservative physical limits of the real vehicle:  
\begin{equation}
\begin{aligned}
\mathcal{A} = \Bigl\{\, 
(a_t, \omega_t) \;\Big|\;
\SI{-2.0}{\meter\per\second\squared} \leq &a_t \leq \SI{2.0}{\meter\per\second\squared}, \\
\SI{-0.5}{\radian\per\second} \leq &\omega_t \leq \SI{0.5}{\radian\per\second}
\,\Bigr\}.
\end{aligned}
\end{equation}

While many works instead use $(a, \gamma)$, i.e. acceleration and steering angle, as in~\cite{maramotti2022tackling}, this can produce infeasible outputs since large instantaneous changes in $\gamma$ may violate steering rate constraints. In contrast, our action representation explicitly respects the dynamic constraints of the steering system.

\textbf{Reward Design}—At each time step $t$, the total reward is
\begin{align}
r_t
=
& R_{\text{dev}}\, h(d_t) \\
+& R_{\text{vel}}\, h\big(\max(0,\, v_t - v_{\max,t})\big) \notag\\
+& R_{\text{progress}}\, \min(v_t,\, v_{\max,t}) \Delta T \notag\\
+& R_{\text{acc}}\, h(a_t)
+ R_{\omega}\, h(\omega_t) \notag\\
+& R_{\text{jerk}}\, h(a_t - a_{t-1})
+ R_{\Delta\omega}\, h(\omega_t - \omega_{t-1}) \notag\\
+& R_{\text{hold}}\, \mathbf{1}\!\{v_{\max,t}\le 0 \wedge v_t>0\},
\label{eq:reward}
\end{align}
where $h(\cdot)$ is a Huber-shaped penalty with scaling
\begin{equation}
h(x)
= \alpha \cdot
\begin{cases}
\frac{1}{2}\, x^2/\delta, & |x|\le \delta,\\[3pt]
|x| - \frac{\delta}{2}, & |x|>\delta,
\end{cases}
\qquad \text{with } \delta=1,\; \alpha=0.25.
\label{eq:huber}
\end{equation}
Here, $d_t$ is the lateral deviation from the reference path and 
$v_{\max,t}=\min\!\big(v_{\text{target,t}},\, v_{\text{centripetal},t}\big)$ 
is the admissible speed limited by the target velocity and the maximum centripetal acceleration $a_{\text{cp}, \max}$. 
The progress term corresponds to the arc-length advance. 
The indicator function $\mathbf{1}\{\cdot\}$ adds a holding penalty if the vehicle 
moves while $v_{\max,t}\le 0$. Each $R_{*}$ is a scalar weight that 
balances the contribution of the respective component.

\begin{figure*}[th!]
    \centering
    \includegraphics[width=\textwidth]{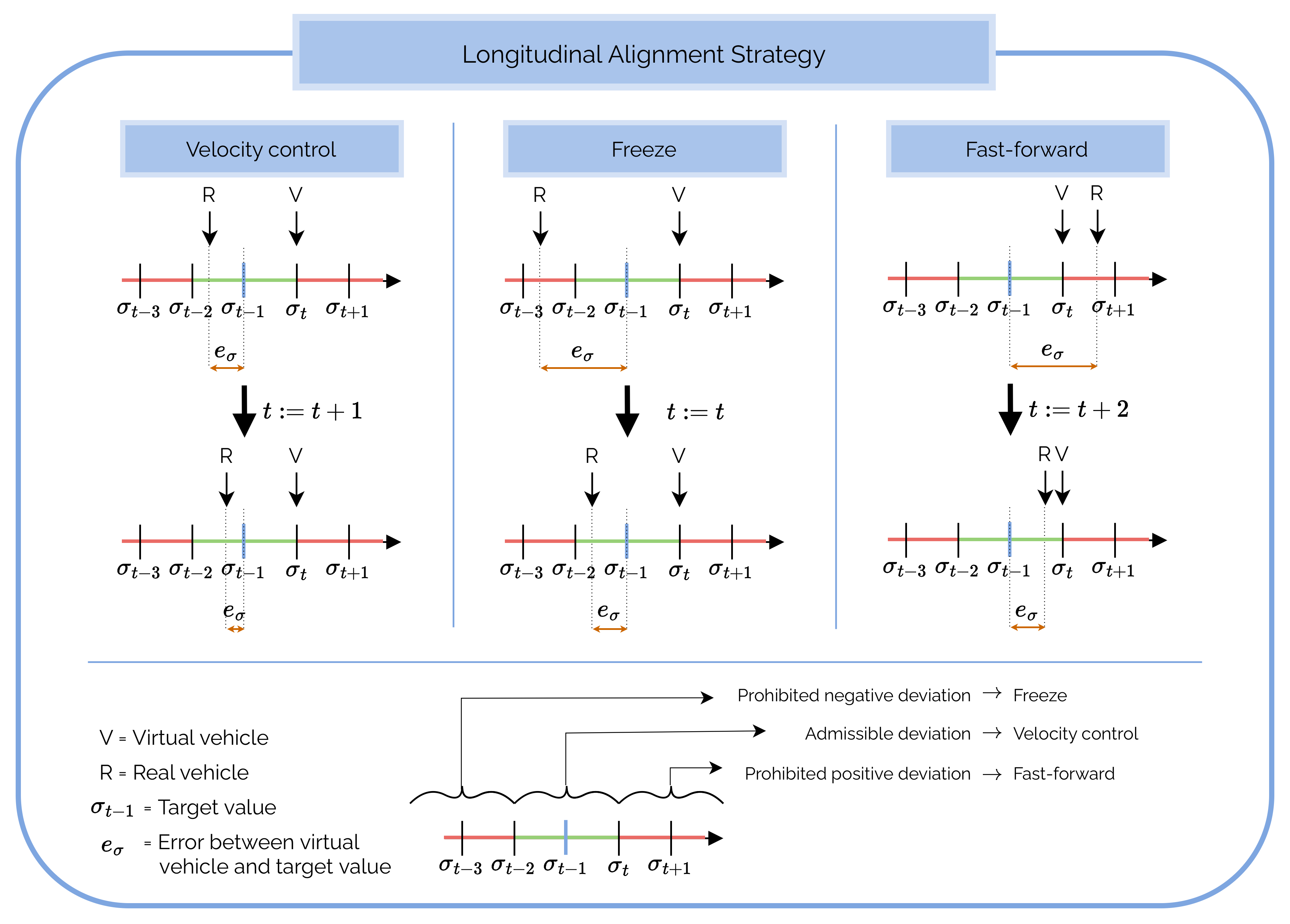}
    \caption{Illustration of three strategies for longitudinal alignment. The virtual vehicle is positioned at $\sigma_{t}$. The real vehicle might have a longitudinal deviation from the desired position $\sigma_{t-1}$. If the position of the real vehicle is within the admissible area (depicted in green) velocity control is applied, e.g. if the real vehicle lacks behind, the velocity is proportionally increased (see the first column of the figure). If the real vehicle is outside the admissible area, two separate strategies are applied: (1) The real vehicle lacks behind. In this case the virtual vehicle does not apply the forward step for the next time step (freeze). Consequently, the real vehicle is moving approximately one time step ahead with respect to the virtual vehicle. (2) The real vehicle is ahead of the virtual vehicle. In this case the virtual vehicle applies two forward steps at once (fast-forward). Consequently, the real vehicle is moving approximately one step back with respect to the virtual vehicle.}
    \label{fig:longitudinal_alignment_strategy}
\end{figure*}
\vspace{-0.2cm}

\subsection{Trajectory Generation}
\label{sec:methods_trajectory_generation}

We derive a trajectory-generating agent by distilling the RL agent described in the previous section. In nominal mode only the first predicted control input is applied during the execution of the agent, while the remaining predictions provide foresight for alignment (see \Cref{sec:methods_alignment}) and can be used for supervision (e.g., collision checking, intention interpretation).  

The agent predicts a trajectory of length $T_H$ consisting of $N_H = T_H / \Delta t$ control pairs (longitudinal acceleration and steering rate). To generate the training dataset, we perform rollouts of the trained RL agent and collect trajectories of length $T_H$ together with their initial states. To improve robustness against distributional shift \cite{lu2023imitation}, Gaussian noise $\mathcal{N}(0,\,0.2)$ is added to the controls. The dataset of size $N_{\text{samples}}$ can be summarized as
\begin{equation}
\mathcal{D} = \big\{ (s_0^{(i)}, \; a_{0:N_H-1}^{(i)}, \; \omega_{0:N_H-1}^{(i)}) \;\big|\; i = 1,\ldots,N_{\text{samples}} \big\},
\end{equation}
where $s_0^{(i)}$ is the initial state, $a_{0:N_H-1}^{(i)}$ the acceleration sequence, and $\omega_{0:N_H-1}^{(i)}$ the steering rate sequence. 

For training, we use an autoregressive approach instead of a one-shot prediction, since future controls depend strongly on previous ones. Predicting each step individually requires $N_H$ inferences, which is costly for large models. Instead, the agent predicts $k$ steps at once, where $N_H$ must be divisible by $k$. The parameter $k$ thus balances accuracy and real-time performance.  

The training objective uses a Euclidean loss on the resulting trajectory poses $(x, y, \theta)$. 
This loss penalizes deviations between the predicted and ground-truth poses along the rollout. 
The weighting factor $\lambda^t$ ($0 \leq \lambda \leq 1$) serves two purposes: it accounts for the 
growing uncertainty over time and increases the relative importance of earlier steps, which directly 
influence the vehicle through applied actions, while later steps primarily serve a supervisory or interpretive role. 
The overall loss is
\begin{equation}
\mathcal{L} = \sum_{t=0}^{N_H-1} \lambda^t \,
\| p_t - \hat{p}_t \|_2^2,
\end{equation}
where $p_t = (x_t, y_t, \theta_t)$ are the ground-truth poses and $\hat{p}_t$ are the corresponding predictions.

\subsection{Trajectory Alignment}
\label{sec:methods_alignment}
Given the trajectory-generating agent from the previous section, we now formulate an alignment strategy between the virtual and real vehicle. The virtual vehicle is cyclically updated in simulation using the kinematic model from \Cref{eq:bicycle_model}, providing the trajectory history, the current state, and the predicted future trajectory for the real vehicle. Along this trajectory, expressed in curvilinear coordinates $(\sigma, d, \Delta \psi)$ (see \Cref{fig:vehicle_modelling}), the real vehicle localizes itself and minimizes both lateral and longitudinal errors. Lateral errors are handled by the Stanley controller, whereas longitudinal alignment requires a dedicated strategy.
To prevent the real vehicle from overtaking the virtual one (which would be undesirable since the predicted trajectory is continuously updated), the real vehicle is designed to follow a reference point that is one time step behind the virtual vehicle. As long as the tracking error remains sufficiently small, the following control law is applied:
\begin{equation}
{}^{\text{R}}a_t = {}^{\text{V}}a_{\tilde{t}} + K_d ({}^{\text{V}}\sigma_{t-1} - {}^{\text{R}}\sigma_{t}) + K_v ({}^{\text{V}}v_{t-1} - {}^{\text{R}}v_{t}),
\label{eq:long_control}
\end{equation}
where the proportional gain for the longitudinal positional feedback is set to $K_d = 1.5$ and the velocity feedback gain to $K_v = 1.0$. The first term, ${}^{\text{V}}a_{\tilde{t}}$, acts as a feed-forward component, applying the virtual vehicle’s acceleration from the closest point on the trajectory to the real vehicle. The second term, $K_d ({}^{\text{V}}\sigma_{t-1} - {}^{\text{R}}\sigma_{t})$, introduces longitudinal positional feedback that minimizes longitudinal deviation, ensuring the real vehicle remains close to the designated reference point. The third term, $K_v ({}^{\text{V}}v_{t-1} - {}^{\text{R}}v_{t})$, provides velocity feedback to mitigate overshoot and enforce smooth convergence of the real vehicle’s speed to that of the virtual reference.

If the real vehicle still overtakes the virtual vehicle, we apply a \emph{fast-forward} step, advancing the virtual vehicle by two steps instead of one. Conversely, if the real vehicle lags behind the lower admissible bound, the virtual vehicle performs a \emph{freeze} step, skipping its next update. These adjustments are feasible because the trajectory generator provides a sequence of control inputs rather than only the next one. The overall longitudinal alignment strategy is illustrated in \Cref{fig:longitudinal_alignment_strategy}.  

In our design, the previous index ${}^{\text{V}}\sigma_{t-1}$ is chosen as reference, and the freeze threshold corresponds to the second-to-last index ${}^{\text{V}}\sigma_{t-2}$. Alternative choices are possible, e.g. interpolating between indices or tightening the freeze/fast-forward bounds. However, windows that are too narrow cause frequent switching, leading to oscillatory velocity corrections and undesirable driving behavior, whereas windows that are too wide reduce reactivity. In our implementation, the real vehicle lags approximately $\Delta T$ behind the virtual vehicle, with a worst case of $2\Delta T$ (i.e., in our case $0.1\,\text{s}$ and $0.2\,\text{s}$, respectively).  

Thus, longitudinal alignment is bidirectional: the velocity control adjusts the real vehicle (${}^{\text{R}}v_t$), while freeze and fast-forward steps adjust the virtual vehicle (${}^{\text{V}}\sigma_t$).


\section{Evaluation}
\label{sec:evaluation}

\subsection{Setting}
\label{sec:evaluation_setting}

We evaluate our approach via a real autonomous vehicle, namely MuCAR-4. The system is deployed on a Nvidia Jetson AGX Orin with ROS2 as middleware. A high-precision inertial navigation system with real-time kinematic global navigation satellite system (GNSS) is used for localization. The control values from the Stanley controller and the longitudinal control are converted to actuator commands by an additional embedded computer. 

\begin{figure}[t!]
    \centering
    \includegraphics[width=\columnwidth]{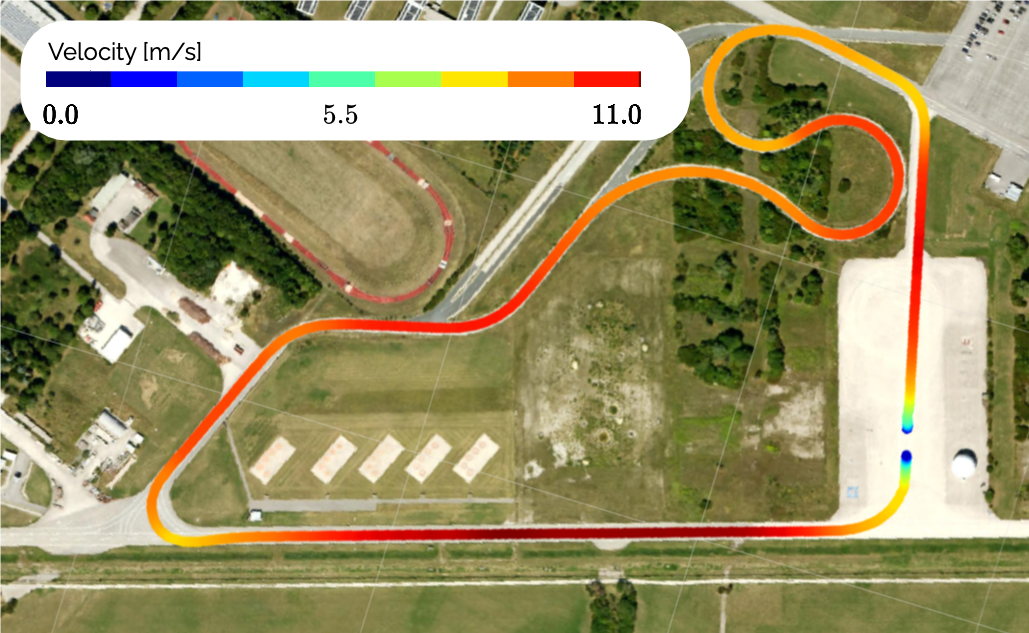}
    \caption{The real-world vehicle test track, measuring \SI{1.9}{\kilo\meter} in length, was driven counter-clockwise at velocities ranging from $0$ to \SI{11}{\meter\per\second} ($\sim$\SI{40}{\kilo\meter\per\hour}).}
    \label{fig:route}
\end{figure}

All experiments were conducted on our university’s closed test area, featuring a fully paved surface. The test route, illustrated in \Cref{fig:route}, covers a total distance of \SI{1.9}{\kilo\meter} and was recorded by driving manually and includes both straight and curved sections representative of urban driving scenarios. Vehicle velocities during the tests ranged from \SI{0}{\meter\per\second} to \SI{11}{\meter\per\second} ($\sim$\SI{40}{\kilo\meter\per\hour}). All positions and trajectories were recorded and processed in the UTM coordinate frame. The goal of the agent is to follow the path recorded by the manual driving and the target speed was specified to also match the speed of the manual execution. 

The trajectory-generating agent consists of a multilayer perceptron (MLP) with four fully connected layers of 512 neurons each, using the SiLU activation function. The network predicts 40 future trajectory points at a temporal resolution of $\Delta T =  \SI{0.1}{\second}$, corresponding to a 4-second prediction horizon. As input, the agent receives the next 80 reference points, sampled equidistantly at 1-meter intervals along the path. The model was implemented in PyTorch and trained using the AdamW optimizer with a learning rate of $5 \times 10^{-4}$ and $\beta = (0.9, 0.9)$. The temporal decay factor in the loss weighting was set to $\lambda = 0.8$, emphasizing earlier steps in the trajectory. Training was performed over 20 epochs with a batch size of 1024 on a dataset of 1,000,000 samples collected from the reinforcement learning agent’s experience. To reduce compounding errors, autoregression was employed with 10 action predictions per inference step, resulting in a total of four sequential inferences to cover the 4-second horizon. The performance of both the trajectory-generating agent and the reinforcement learning agent is not analyzed, as it is independent of the proposed sim-to-real approach.

\subsection{Results}
\label{sec:evaluation_results}

\begin{figure*}[ht!]
    \centering
    \includegraphics[width=\textwidth]{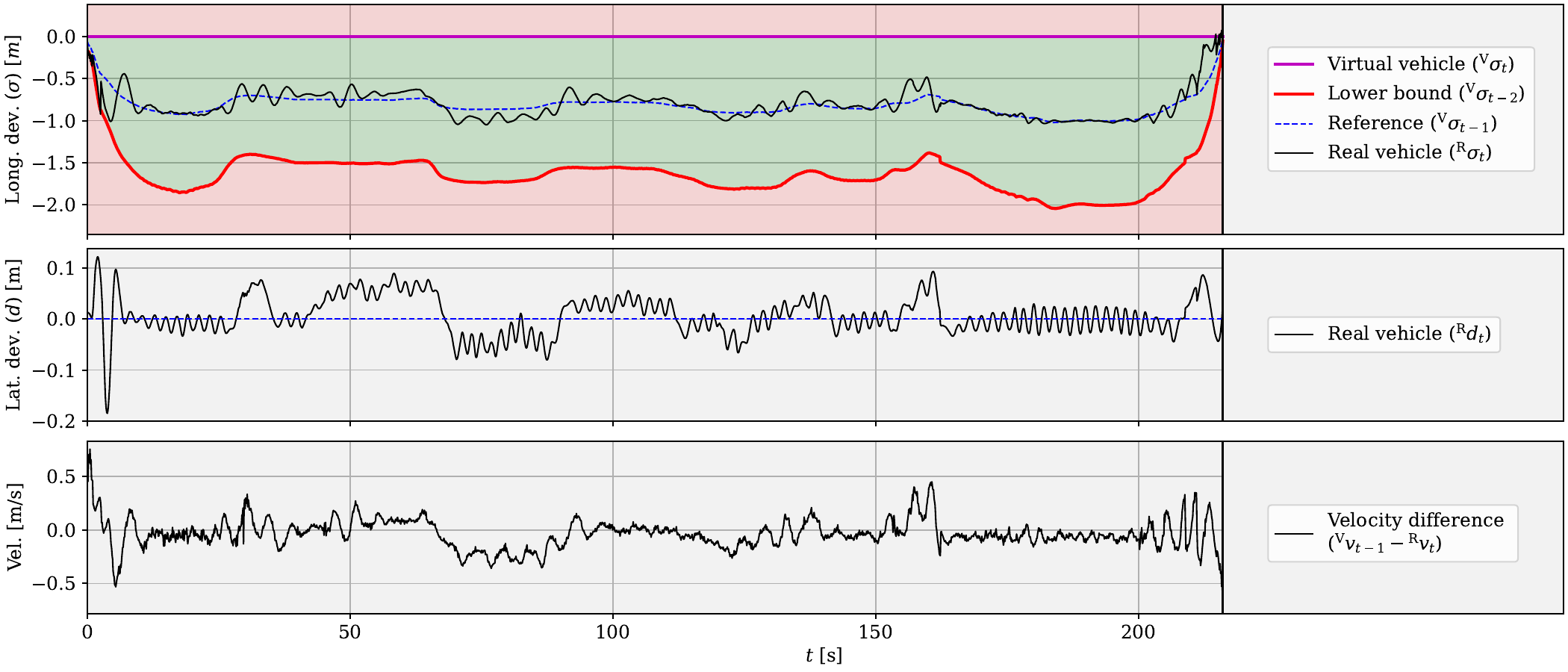}
    \caption{
    Results from the real-vehicle experiment. A digital version of the figure is recommended for improved visibility of details. 
    \textbf{Top}: longitudinal alignment, with the green area indicating the admissible deviation of the real vehicle with respect to the virtual vehicle, and the blue dashed line denoting the desired reference. The longitudinal position of the real vehicle (${}^{\text{R}}\sigma_{t}$), the lower bound (${}^{\text{V}}\sigma_{t-2}$), and reference (${}^{\text{V}}\sigma_{t-1}$) are shown relative to the virtual vehicle. 
    \textbf{Center}: lateral deviation of the real vehicle with respect to the virtual trajectory. 
    \textbf{Bottom}: velocity difference between the desired and the real vehicle's velocity.
    }
    \label{fig:eval_real_vehicle}
\end{figure*}
\vspace{-0.2cm}

The results of our experiment are depicted in \Cref{fig:eval_real_vehicle}. The longitudinal deviation along the entire track remained close to the reference position, corresponding to one time step behind the virtual vehicle. The largest deviations occurred at low velocities, i.e., at the beginning and end of the track, which likely correlate with higher acceleration magnitudes that are more difficult to control. Additionally, low velocities may amplify measurement noise, as the wheel-speed sensors used for estimating the vehicle’s velocity can become less reliable in this regime. During the stopping phase (\SIrange{217}{220}{\second}), shortly before standstill, spikes and abrupt changes in the longitudinal deviation can be observed. These result from \emph{fast-forward steps}, where the real vehicle temporarily exceeded the virtual vehicle, leading to two forward updates being applied. Similarly, at the start of motion (\SIrange{0.0}{3.0}{\second}), \emph{freeze steps} can be observed, where the virtual vehicle did not advance for several time steps to allow the real vehicle to catch up. This design choice prevents the virtual vehicle from falling below the lower bound and ensures synchronized behavior. The mean absolute longitudinal error is \SI{6.8}{\centi\meter}, while the maximum error of \SI{50.0}{\centi\meter} occurred during the start of the track. 

The center plot of \Cref{fig:eval_real_vehicle} shows the lateral error of the real vehicle with respect to the virtual trajectory. Similar to the longitudinal case, the largest deviation is observed at the start, with a maximum of \SI{18.5}{\centi\meter}. After the initial phase, the control error remains below \SI{10}{\centi\meter}, even at velocities up to \SI{11}{\meter\per\second} ($\sim$\SI{40}{\kilo\meter\per\hour}), demonstrating the robustness and applicability of the proposed approach. The mean absolute lateral deviation is \SI{2.9}{\centi\meter}. Nevertheless, we believe that lateral accuracy could be further improved using an enhanced controller, such as the Enhanced Stanley controller proposed in \cite{seiffer2023pragmatic}, which achieved accuracies below \SI{8}{\centi\meter} for feasible desired trajectories and reduced oscillatory control behavior. 

Finally, the bottom plot in \Cref{fig:eval_real_vehicle} shows the velocity difference between the real and virtual vehicle. This deviation originates from the longitudinal control law described in \Cref{eq:long_control} and reflects both model inaccuracies and the real vehicle’s dynamic response. The maximum velocity error is \SI{0.71}{\meter\per\second}, occurring at the start of motion. Apart from this initial transient, the error remains below \SI{0.5}{\meter\per\second}, with a mean absolute velocity error of \SI{0.11}{\meter\per\second}.

To conclude, it should be noted that conducting experiments without longitudinal control is not feasible. Without it, the virtual vehicle would quickly move ahead, which is impractical since perception data is only available at the real vehicle. Conversely, if the real vehicle were to move ahead, no valid reference path would be available, making the experiment equally infeasible.


\subsection{Discussion}
\label{sec:discussion}

The results in \Cref{sec:evaluation_results} demonstrate that our alignment strategy maintains small tracking errors, making it suitable for urban driving scenarios. Further reduction of the virtual-to-real discrepancy could be achieved by employing more advanced control methods, such as the Enhanced Stanley controller \cite{seiffer2023pragmatic} or a model predictive controller (MPC) \cite{bienemann2024perception}. In simulation, we relied on a simplified bicycle model, which requires the control strategy to adapt accordingly. Although our experiments show that this level of fidelity is sufficient, performance could be improved by incorporating more accurate vehicle dynamics models.  

For specific driving conditions, such as drifting, the bicycle model becomes insufficient, and the simple alignment control may fail if a significant dynamics mismatch arises between the real and virtual vehicles. In such extreme cases, two mitigation strategies are possible: (1) improve the simulation model to better match the real vehicle’s dynamics—although this is often impractical, as it would require perception of complex physical parameters such as tire condition, mass distribution, and surface properties—or (2) reset the alignment when the lateral deviation exceeds a threshold. While the latter is not optimal for control continuity, it effectively prevents hazardous behavior caused by severe virtual-to-real mismatch.


\section{Conclusion}
\label{sec:conclusion}

In this paper, we present a novel strategy to deploy a reinforcement learning-trained agent to a real autonomous vehicle in a zero-shot manner by aligning a virtual and a real vehicle. The proposed alignment approach effectively minimizes both longitudinal and lateral errors, enabling stable real-world execution without additional fine-tuning. Our results demonstrate that the deviation between the real and virtual vehicle remains small, indicating that this straightforward zero-shot sim-to-real transfer strategy can be generalized to other robotic domains beyond autonomous driving. An important implication of our approach is its robustness to variations in real-world dynamics, such as changes in tire properties, road friction, or weather conditions, which typically degrade the performance of sim-to-real methods. Future work will focus on enhancing control robustness and evaluating the system under more challenging conditions, including off-road environments and higher velocities.


\bibliographystyle{IEEEtran}  \balance 
\bibliography{macros/IEEEabrv,macros/additional_abrv,macros/et_al,ref}

\end{document}